\documentclass{article}
\usepackage{times}
\usepackage{soul}
\usepackage{url}
\usepackage[utf8]{inputenc}
\usepackage[small]{caption}
\usepackage{graphicx}
\usepackage{amsmath}
\usepackage{booktabs}
\usepackage{algorithm}
\usepackage{caption}
\usepackage{subcaption}
\usepackage{algpseudocode}
\usepackage{amsmath}
\usepackage{amssymb}

\title{Zero-shot Domain Adaptation\\ Based on Attribute Information}

\author{
Masato Ishii$^{1,2,3}$
\and
Takashi Takenouchi$^{2,4}$\and
\vspace{2mm}
Masashi Sugiyama$^{2,1}$\\
$^1$ The University of Tokyo\\
$^2$ Center for Advanced Intelligence Project, RIKEN\\
$^3$ Data Science Research Laboratories, NEC\\
$^4$ Future University Hakodate\\
}
\date{}

\begin{document}

\maketitle

\begin{abstract}
In this paper, we propose a novel domain adaptation method that can be applied without target data. We consider the situation where domain shift is caused by a prior change of a specific factor and assume that we know how the prior changes between source and target domains. We call this factor an attribute, and reformulate the domain adaptation problem to utilize the attribute prior instead of target data. In our method, the source data are reweighted with the sample-wise weight estimated by the attribute prior and the data themselves so that they are useful in the target domain. We theoretically reveal that our method provides more precise estimation of sample-wise transferability than a straightforward attribute-based reweighting approach. Experimental results with both toy datasets and benchmark datasets show that our method can perform well, though it does not use any target data. 
\end{abstract}

\section{Introduction}

\begin{figure}
\centering
\subcaptionbox{Video A}{\includegraphics[scale=0.85]{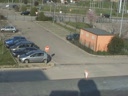}}
\subcaptionbox{Video B}{\includegraphics[scale=0.85]{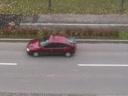}}
\caption{Example images of surveillance videos. Since the position and pose of the surveillance camera is different, the appearance of the vehicle is somewhat different between two videos.}
\label{fig:exm_video}
\end{figure}

In many algorithms for supervised learning, it is assumed that training data are obtained from the same distribution as that of test data \cite{Hastie2009}. Unfortunately, this assumption is often violated in practical applications. For example, Fig. \ref{fig:exm_video} shows images of two different surveillance videos that are obtained from Video Surveillance Online Repository \cite{Vezzani2010}. Suppose we want to recognize vehicles from these videos. Since the position and pose of the camera are different, the appearance of the vehicle is somewhat different between two videos. Due to this difference, even if we train a highly accurate classifier on video A, it may work poorly on video B. Such discrepancy has recently become a major problem in pattern recognition, because it is often difficult to obtain training data that are sufficiently similar to the test data. To deal with this problem, domain adaptation techniques have been proposed. 

Given two datasets, called source and target data, domain adaptation aims to adapt source domain data to the target domain data so that distributions of both datasets are matched \cite{Csurka2017}. By applying domain adaptation, classifiers trained on the adapted source data can achieve high accuracy on the target data. Since the discrepancy between two distributions is measured based on observed data, we need a sufficient number of data in each dataset to estimate the distributional discrepancy accurately. However, due to the motivation of the domain adaptation, obtaining a large number of target data is often hard, which limits the application of domain adaptation methods to practical cases.

In this work, we consider the most extreme case in which we cannot obtain any target data, called zero-shot domain adaptation. A few recent studies \cite{Yang2015,Peng2018} have tackled this challenging problem, but they require additional data such as multiple source datasets \cite{Yang2015} or target data of another task \cite{Peng2018} that are not easy to obtain in practice. 
In this paper, we propose a novel method of zero-shot domain adaptation that would be more suitable for practical cases. We assume that we have prior knowledge about what factor causes the difference in distributions between source and target data. For example, in Fig. \ref{fig:exm_video}, the shooting angle for vehicles can be considered as a major factor that causes the appearance change between videos. Other examples include gender information in an age estimation task from facial images and the azimuth of captured objects in an object recognition task, both of which are examined in our experiments. 

We call such a factor an attribute, and assume that we can only obtain attribute priors at the target domain instead of the target data. We then reformulate the domain adaptation problem so that we can conduct adaptation based only on attribute priors. In addition, we clarify requirements for the attribute to be useful in domain adaptation, and reveal that our method provides more precise estimation of sample-wise transferability than the straightforward attribute-based reweighting approach. Experimental results with both toy datasets and benchmark datasets validate the advantage of our method, even though it does not use any target data. 

We explain our setting by using vehicle recognition from surveillance videos as an example. In this task, input data and labels are cropped video frames and vehicle types, respectively. Suppose that we have already constructed training datasets from existing surveillance cameras and want to transfer those datasets to a classifier for a new surveillance camera. If the new camera is not installed yet, we cannot obtain any target videos, therefore, we cannot apply a standard domain adaptation method nor evaluate how much data can be transfered via domain adaptation. But, if where and how the new camera will be installed have been already determined, we can estimate the shooting angle for the target vehicle. Since the shooting angle is a major factor that causes the appearance change of vehicles, we can consider the shooting angle as an attribute. In this case, we calculate it for each sample at the source domain and also estimate how often the vehicle will be captured with a certain shooting angle at the target domain by using the information about the pose and position of a camera. As shown in the above example, the assumption about attribute information in our method is sufficiently practical, and we believe that our method can be applied in many practical applications, especially for computer vision tasks. 

\section{Problem formulation and related works}

Recent domain adaptation methods \cite{Ganin2016,Tzeng2017,Peng2018} often adopt deep neural networks (DNNs) to embed both-domain data into a common feature space in which the distributions of both data are matched. But, due to the ``data-hungry" property of DNN, this approach requires a relatively large number of data. Since we tackle the ``zero-shot" scenario in which we cannot obtain any target data, we utilize a different approach in this work, that is the instance-weight based approach \cite{Huang2007,Sugiyama2008,Kanamori2009}. In this approach, domain adaptation is achieved by assigning an instance weight for each sample in the source data. 

We briefly show the problem setting of the domain adaptation and how to solve it by the instance-weight based approach. 
Let us consider a supervised classification task, and let $x \in \mathbb{R}^m, y \in C$ and $d \in \{{\rm S}, {\rm T}\}$ denote input data, labels, and domains, respectively. Here, $m$ is the dimensionality of the input data, $C$ is the set of the class candidates, and $\{{\rm S}, {\rm T}\}$ represent the source and target domains, respectively. Note that we treat $d$ as a random variable. We assume that the joint distributions of $(x,y)$ are different between domains, which means $p(x,y|d={\rm S}) \neq p(x,y|d={\rm T})$. Given labeled source data $\mathcal{D}_S = \{(x^{{\rm S}}_i,y^{{\rm S}}_i)\} \sim p(x,y|d={\rm S})$ and unlabeled target data $\mathcal{D}_T = \{x^{{\rm T}}_i\} \sim p(x|d={\rm T})$, our goal is to train a model $f: \mathbb{R}^m \rightarrow C$ that can accurately predict labels for input data at the target domain. More specifically, supposing $f$ is parameterized by $\theta$, we want to obtain the optimal $\theta$ that minimizes the target risk defined as
\begin{eqnarray}
R_\mathrm{T}(\theta) \!\!\!&=&\!\!\! \sum_{y \in C} \int l(x,y,\theta) p(x,y|d=\mathrm{T}) {\rm d}x, 
\label{eq:risk_T}
\end{eqnarray}
where $l(x,y,\theta)$ is a loss when $y$ is predicted by $f$ with $\theta$ at $x$. 

Since the target data are not labeled, we cannot directly estimate the risk in Eq. (\ref{eq:risk_T}) by empirical approximation. Instead, we try to use the source data to estimate it. The target risk can be related to the source risk with instance weights as: 
\begin{eqnarray}
R_\mathrm{T}(\theta) \!\!\!&=&\!\!\! \sum_{y \in C} \int w(x,y) l(x,y,\theta) p(x,y|d=\mathrm{S}) {\rm d}x,
\label{eq:weight_risk}
\end{eqnarray}
where $w(x,y) = \frac{p(x,y|d=\mathrm{T})}{p(x,y|d=\mathrm{S})}$ is an instance weight for the corresponding data $(x,y)$. By assuming covariate shift \cite{Shimodaira2000}, that means $p(y|x)$ is common in the source and target domains, we can simplify the weight as follows
\begin{eqnarray}
\frac{p(x,y|d=\mathrm{T})}{p(x,y|d=\mathrm{S})} &=& \frac{p(y|x,d=\mathrm{T})}{p(y|x,d=\mathrm{S})} \frac{p(x|d=\mathrm{T})}{p(x|d=\mathrm{S})} \nonumber \\
 &=& \frac{p(x|d=\mathrm{T})}{p(x|d=\mathrm{S})} = w(x).
\label{eq:ratio}
\end{eqnarray}

Equation (\ref{eq:weight_risk}) indicates that we can obtain the optimal $\theta$ by minimizing the weighted source risk. Therefore, many existing instance-weight based methods \cite{Huang2007,Sugiyama2008,Kanamori2009} basically try to accurately estimate the weight defined in Eq. (\ref{eq:ratio}). When we estimate the weight, we assume that the weight is always finite. Once we obtain the weight for each sample in the source data, we can calculate the empirically approximated risk $\hat{R_{{\rm T}}}(\theta)$ as:
\begin{eqnarray}
\hat{R_{{\rm T}}}(\theta) = \frac{1}{|\mathcal{D}_S|} \sum_{(x_i, y_i) \in \mathcal{D}_S} \hat{w}(x_i) l(x_i, y_i, \theta),
\end{eqnarray}
where $\hat{w}(x_i)$ is the estimated weight for $(x_i,y_i)$. By minimizing this empirical risk, we can estimate the optimal $\theta$.

In our zero-shot scenario, the standard instance-weight based approach cannot be adopted, because they require target data as well as source data to estimate the weight. Therefore, the main problem in our scenario is how to estimate the weight without target data. We will show that it can be solved by utilizing the attribute information instead of the unavailable target data.

\section{Zero-shot domain adaptation based on attribute information}

We assume that we can obtain attribute information at both the source and target domains that is a major factor for the discrepancy between the data distributions. More specifically, at the source domain, attribute $z$ for each sample is also given in addition to $(x,y)$, and at the target domain, we cannot obtain any data or attributes as well, but only the probability distribution of attributes $p(z|d=\mathrm{T})$ is given. To make our formulation simple, we assume a single categorical attribute, but our method can be extended to multivariate or continuous attributes in a straightforward way.

\subsection{How to calculate instance weights}
First, we transform the probability density ratio in Eq. (\ref{eq:ratio}).
Since we do not have any information about the domain prior $p(d)$ especially for the target domain, we assumed a uniform distribution ($p(d=\mathrm{S})=p(d=\mathrm{T})$) that is often used as a non-informative prior. By using this assumption and Bayes' theorem, we obtain the following equation:
\begin{eqnarray}
w(x) \!=\! \frac{p(x|d\!=\!\mathrm{T})}{p(x|d\!=\!\mathrm{S})} 
 \!=\! \frac{p(d\!=\!\mathrm{T}|x)}{p(d\!=\!\mathrm{S}|x)} \frac{p(d\!=\!\mathrm{S})}{p(d\!=\!\mathrm{T})} \!=\! \frac{p(d\!=\!\mathrm{T}|x)}{p(d\!=\!\mathrm{S}|x)}.\!\!\!
\label{eq:w_mid}
\end{eqnarray}
Then, based on the attribute information, we approximate $p(d|x)$ as follows:
\begin{eqnarray}
p(d|x) \approx \sum_z p(d|z) p(z|x).
\label{eq:approx}
\end{eqnarray}
We will discuss what condition is required for the approximation in Eq. (\ref{eq:approx}) in the next subsection. Substituting Eq. (\ref{eq:approx}) into Eq. (\ref{eq:w_mid}), we obtain 
\begin{eqnarray}
w(x) = \frac{\sum_z p(d=\mathrm{T}|z) p(z|x)}{\sum_z p(d=\mathrm{S}|z) p(z|x)}.
\label{eq:weight}
\end{eqnarray}
By adopting the approximation in Eq. (\ref{eq:approx}), we can calculate $w(x)$ by estimating $p(d|z)$ and $p(z|x)$. It means that we do not need the target data, because $p(d|z)$ can be estimated from the given information about the attributes, and $p(z|x)$ that does not depend on domains can be estimated from the source data. This is the key trick of our method. 

Since we assume that $p(z|d)$ is given and $p(d={\rm S})=p(d={\rm T})$, $p(d|z)$ can be calculated by using Bayes' theorem as follows: 
\begin{eqnarray}
\label{eq:pdzT}
p(d=\mathrm{T}|z) &=& \frac{p(z|d=\mathrm{T})}{p(z|d=\mathrm{S})+p(z|d=\mathrm{T})}, \\
p(d=\mathrm{S}|z) &=& 1- p(d=\mathrm{T}|z)
\label{eq:pdzS}
\end{eqnarray}
For the estimation of $p(z|x)$, we adopt the $k$-nearest neighbor method which is the simplest method for the posterior estimation: given $x$, we search $k$ nearest samples from the source data and extract the corresponding attributes. Since we assumed that the attributes are categorical, we calculate the proportion of each attribute class within the extracted attributes. If the attribute is continuous, we may use kernel density estimation. 

\begin{algorithm}[t]
\caption{Zero-shot domain adaptation}
\begin{algorithmic}[1]
\Require{Source data $(x,y,z) \sim p(x,y,z|d=S)$ is given}
\Require{Target attribute information $p(z|d=T)$ is given}
\Require{Equation (\ref{eq:approx}) and $p(d=S)=p(d=T)$ hold}
\State{Calculate $p(d|z)$ by Eq. (\ref{eq:pdzT}) and (\ref{eq:pdzS})}
\State{Estimate $p(z|x)$ with the source data ($k$-NN method is used in this paper)}
\State{Calculate $w(x)$ by Eq. (\ref{eq:weight}) using $p(d|z)$ and $p(z|x)$}
\State{\Return{$w(x)$}}
\end{algorithmic}
\end{algorithm}

\subsection{Requirements for the attribute information}

The most important assumption in our method is Eq. (\ref{eq:approx}). In this subsection, we clarify requirements for this approximation. Since $p(d|x)$ equals $\sum_z p(d|x,z)p(z|x)$, we need the following approximation to have Eq. (\ref{eq:approx}): 
\begin{eqnarray}
p(d|x,z) \approx p(d|z).
\label{eq:pdxz}
\end{eqnarray}
By multiplying $p(x|z)$ to both sides of Eq. (\ref{eq:pdxz}), we can obtain $p(d,x|z) \approx p(d|z) p(x|d)$. Therefore, this approximation assumes that $x$ and $d$ are conditionally independent given $z$. 

We show another aspect of this approximation. By using Bayes' theorem, the left-hand side of Eq. (\ref{eq:pdxz}) can be transformed as follows: 
\begin{eqnarray}
p(d|x,z) = \frac{p(x,z|d)p(d)}{p(x,z)}
 \!\!\!\!&=&\!\!\!\! \frac{p(x|z,d)p(z|d)p(d)}{p(x|z)p(z)} \nonumber \\
 \!\!\!\!&=&\!\!\!\! \frac{p(x|z,d)}{p(x|z)} p(d|z).
\label{eq:mid}
\end{eqnarray}
By substituting Eq. (\ref{eq:mid}) into Eq. (\ref{eq:pdxz}), we obtain
\begin{eqnarray}
p(x|z,d) \approx p(x|z).
\label{eq:assumption}
\end{eqnarray}
This equation indicates that, given a certain $z$, the probability distribution of $x$ is common between domains. Since marginal probability density $p(x|d) = \sum_z p(x|z,d) p(z|d)$ is different between the source and target domains while $p(x|z)$ is common, only the attribute prior given a domain $p(z|d)$ is different between domains. Therefore, the approximation in Eq. (\ref{eq:approx}) corresponds to the {\it latent prior change assumption} that is adopted in some existing works \cite{Storkey2007,Hu2018}.

\subsection{Characteristics of the proposed method}

We clarify some characteristics of our method. First, we take two special cases to explain how our method works, and after that we show how our method is different from the straightforward attribute-based instance weighting. 

If the attribute prior is identical between the source and target domains, that means $p(z|d\!=\!\mathrm{S})\!=\!p(z|d\!=\!\mathrm{T})$, $p(d|z)$ in Eqs. (\ref{eq:pdzT}) and (\ref{eq:pdzS}) are always $0.5$ regardless of the value of $z$. This results in $w(x) \!=\! 1$, which indicates that the source data have already been adapted to the target data and we do not need to conduct domain adaptation. This is natural behavior, because we assumed that only the attribute prior changes between domains as noted in the previous subsection.

If $p(z|x)$ is the delta function $\delta (z=z^*)$ where $z^*$ is the attribute value that corresponds to given sample $x$, $w(x)$ in Eq. (\ref{eq:weight}) can be simplified as follows:
\begin{eqnarray}
w(x) = \frac{p(d=\mathrm{T}|z=z^*)}{p(d=\mathrm{S}|z=z^*)} = \frac{p(z=z^*|d=\mathrm{T})}{p(z=z^*|d=\mathrm{S})}.
\end{eqnarray}
This means that the weight is determined based on only attribute information and not on data. It corresponds to the straightforward approach for attribute-based instance weighting. If we define the weight as $w(x,y,z)=\frac{p(x,y,z|d=\mathrm{T})}{p(x,y,z|d=\mathrm{S})}$ and assume $p(x,y|z,d=\mathrm{S})=p(x,y|z,d=\mathrm{T})$ that is somewhat a stronger assumption in Eq. (\ref{eq:assumption}), we can derive the above instance weight as follows:
\begin{eqnarray}
w(x,y,z) \!=\! \frac{p(x,y,z|d=\mathrm{T})}{p(x,y,z|d=\mathrm{S})} 
\!\!\!\!\!&=&\!\!\!\!\! \frac{p(x,y|z,d=\mathrm{T})p(z|d=\mathrm{T})}{p(x,y|z,d=\mathrm{S})p(z|d=\mathrm{S})} \nonumber \\
&=&\!\!\!\!\! \frac{p(z|d=\mathrm{T})}{p(z|d=\mathrm{S})}.
\label{eq:straight}
\end{eqnarray}
As shown above, our method includes the straightforward attribute-based method as a special case. In other cases, that mean $p(z|x)$ is not a delta function, our method behaves differently compared with the straightforward method. 

\begin{figure}[t]
\centering
\subcaptionbox{$p(x|z)$ and estimated weight \label{fig:exmA_w}}{\includegraphics[scale=0.26]{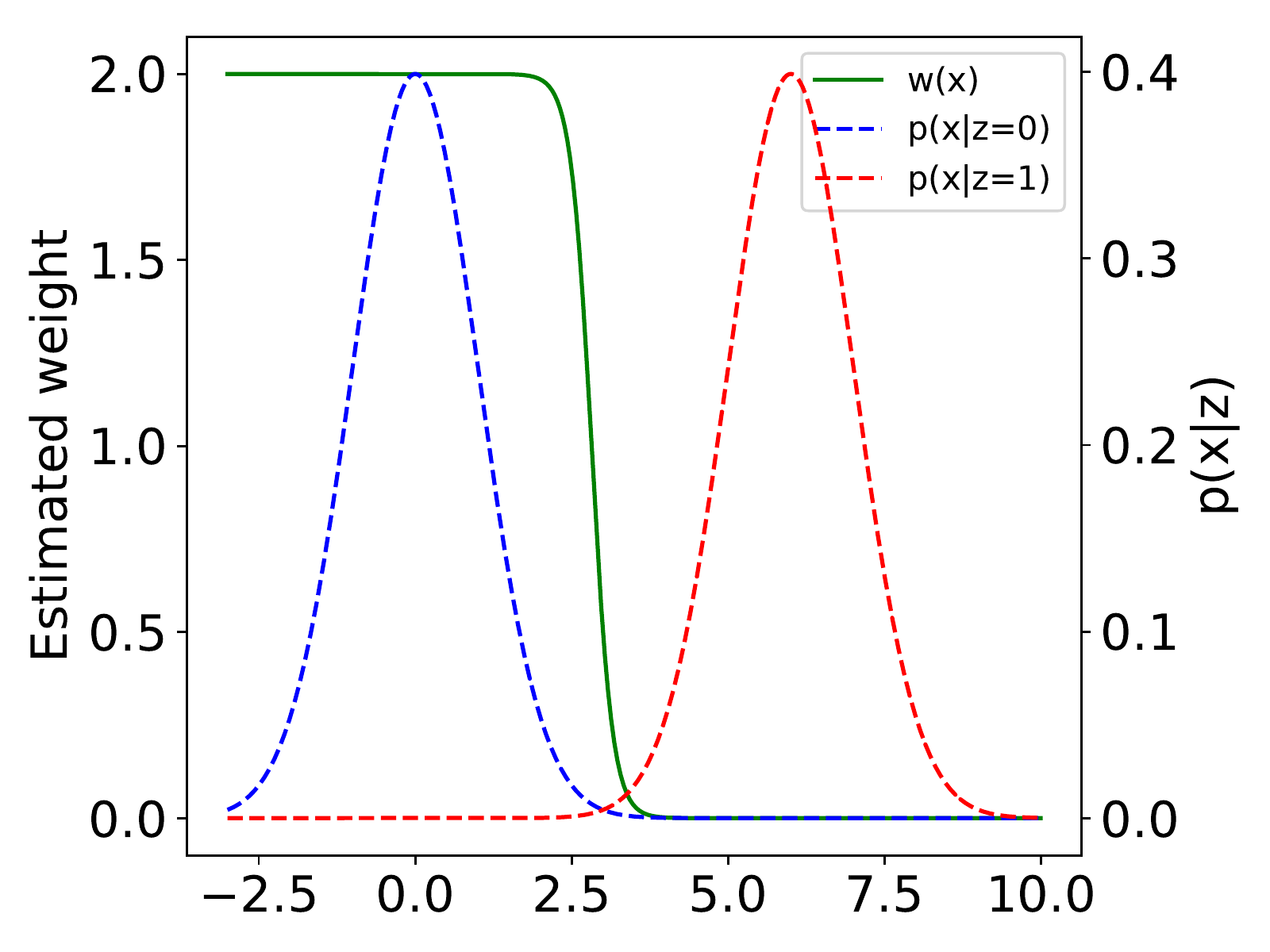}}
\subcaptionbox{Histogram of the weight for each attribute class \label{fig:exmA_wh}}{\includegraphics[scale=0.26]{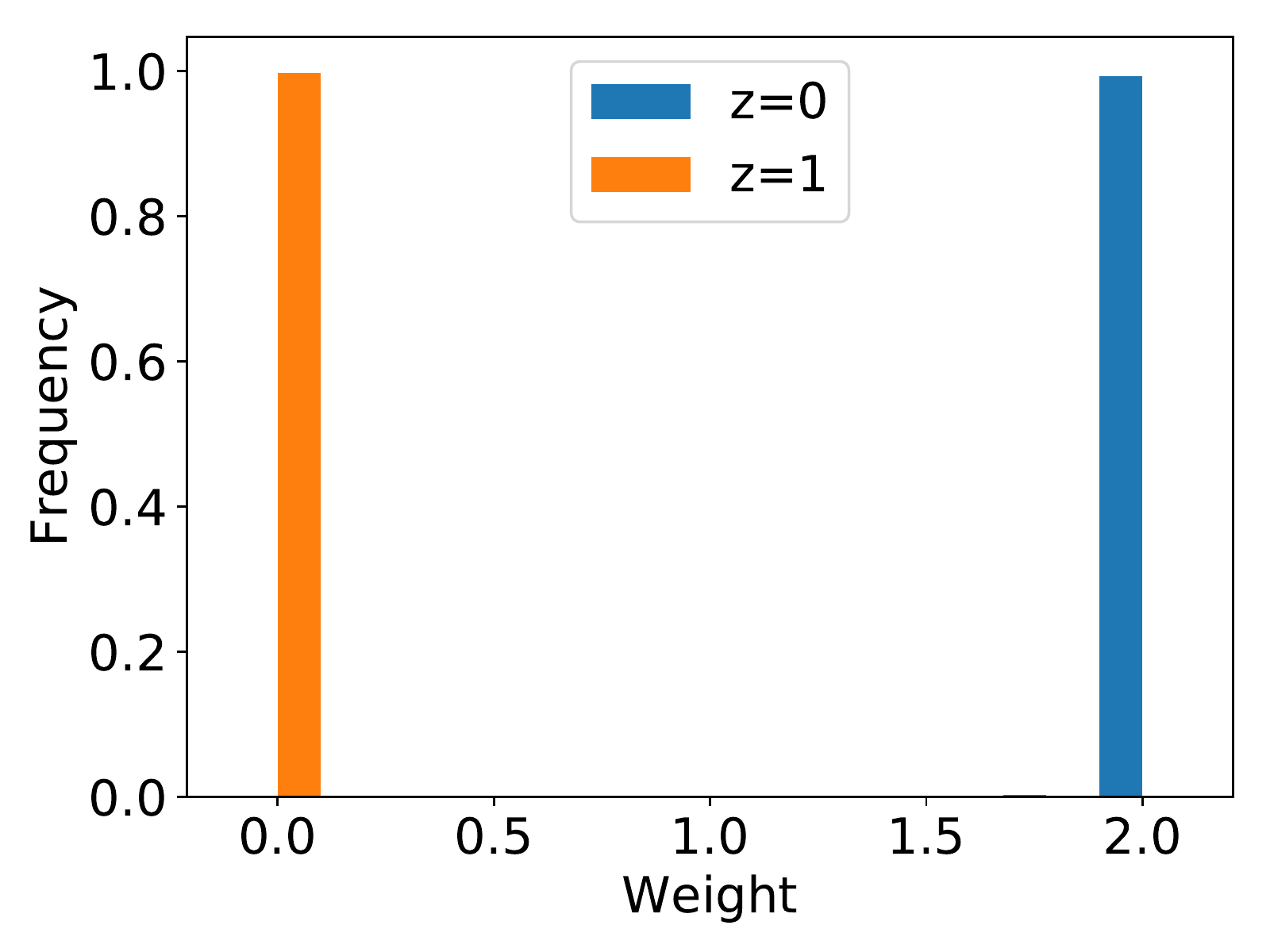}}
\caption{One-dimensional example when the overlap between $p(x|z=0)$ and $p(x|z=1)$ is small.}
\label{fig:exmA}

\centering
\subcaptionbox{$p(x|z)$ and estimated weight \label{fig:exmB_w}}{\includegraphics[scale=0.26]{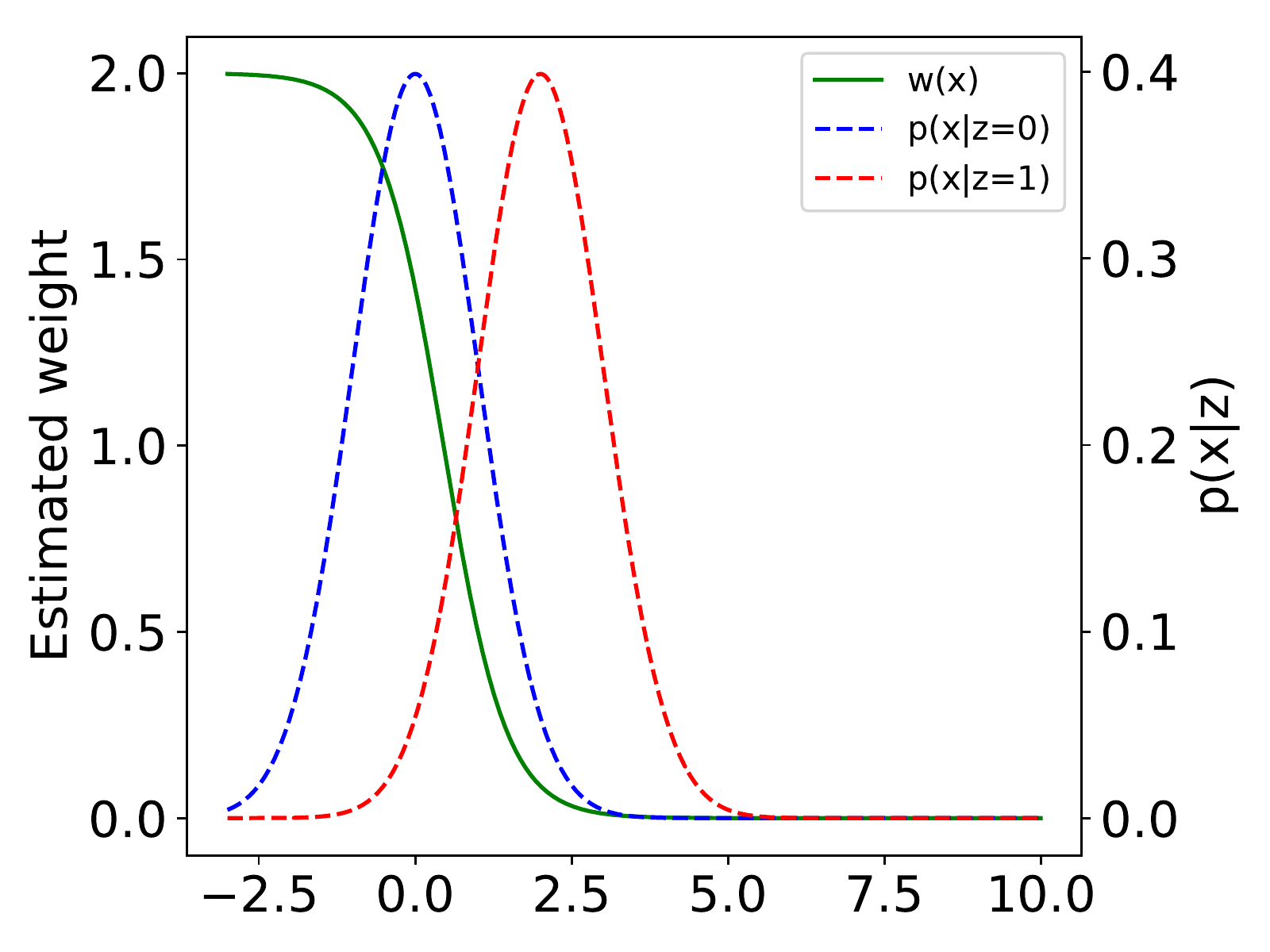}}
\subcaptionbox{Histogram of the weight for each attribute class \label{fig:exmB_wh}}{\includegraphics[scale=0.26]{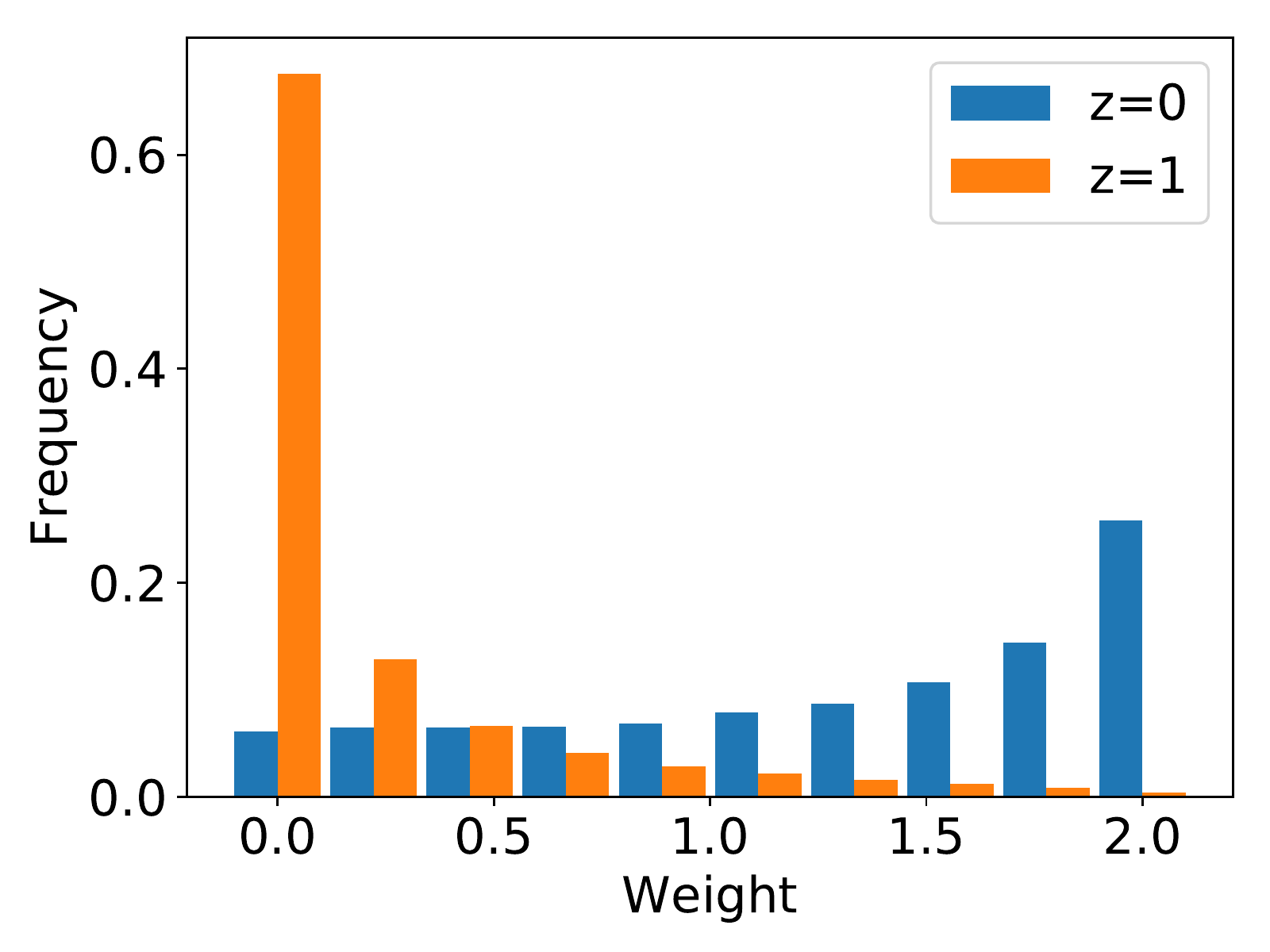}}
\caption{One-dimensional example when the overlap between $p(x|z=0)$ and $p(x|z=1)$ is large.}
\label{fig:exmB}
\end{figure}

Let us illustrate the behavior of our method using a simple example. 
Suppose there are only two attribute classes $z \in \{0, 1\}$ that have one-dimensional Gaussian distributions with different means as shown in Fig. \ref{fig:exmA_w}. At the source domain, $[ p(z\!=\!0|d\!=\!\mathrm{S}), p(z\!=\!1|d\!=\!\mathrm{S})]$ is set to $[0.5, 0.5]$, while it is set to $[1.0, 0.0]$ at the target domain. In this case, the weight estimated in the straightforward method (Eq. (\ref{eq:straight})) leads to a simple delta function, that is $w(x,y,z) \!=\! 2\cdot\delta(z\!=\!0)$. In contrast, the weight in our method (Eq. (\ref{eq:weight})) behaves differently according to the amount of overlap between $p(x|z\!=\!0)$ and $p(x|z\!=\!1)$. Figure \ref{fig:exmA} shows the case in which the overlap is quite small. The weight function $w(x)$ becomes almost the same as a step function over $x$ as shown in Fig. \ref{fig:exmA_w}. As a result, the weight over $z$ becomes the delta function that is the same as that in the straightforward method as shown in Fig. \ref{fig:exmA_wh}. In contrast, when the overlap is large, our method shows somewhat different behavior as presented in Fig. \ref{fig:exmB}. In this case, $w(x)$ becomes a smoother function compared with the previous case as shown in Fig. \ref{fig:exmB_w}. It leads to non-zero weights for the samples with $z\!=\!1$ as shown in Fig. \ref{fig:exmB_wh}, which means that we can transfer these samples even though the samples with $z\!=\!1$ do not appear at the target domain. This characteristic is not available in the straightforward method, because it focuses only on the attribute to estimate the weight. On the other hand, our method utilizes the information of $p(z|x)$, which results in smoother weights that can transfer the source data more efficiently.

\section{Experiments}

In this section, we show the experimental results with both toy datasets and benchmark datasets. 

\subsection{Experiments with toy dataset}

We conducted experiments with a 2-dimensional toy dataset for binary classification, In this dataset, the first feature $x_0$ stemmed from a Gaussian mixture model (GMM) that has five centroids ($-0.75\pi,\ -0.5\pi,\ 0.0,\ 0.5\pi,\ 0.75\pi$) with common standard deviation $\sigma = 0.2\pi$, and the second feature $x_1$ stemmed from the uniform distribution from $-2.0$ to $2.0$. For each sample, the index of the corresponding centroid was treated as attribute $z \in \{0,1,2,3,4\}$. The mixing ratio of GMM was set differently for the source and target domains as shown in Table \ref{tab:GMM}. To change the difficulty of domain adaptation, we constructed three datasets (Datasets A--C) by changing the discrepancy of the ratios between the domains. The posterior $p(y|x)$ is determined by $p(y|x) = \frac{1}{1 + {\rm exp}\left(-5.0 (x_1 - \sin x_0)\right)}$.

\begin{figure}[t]
\centering
\includegraphics[scale=0.3,viewport=0 60 720 480]{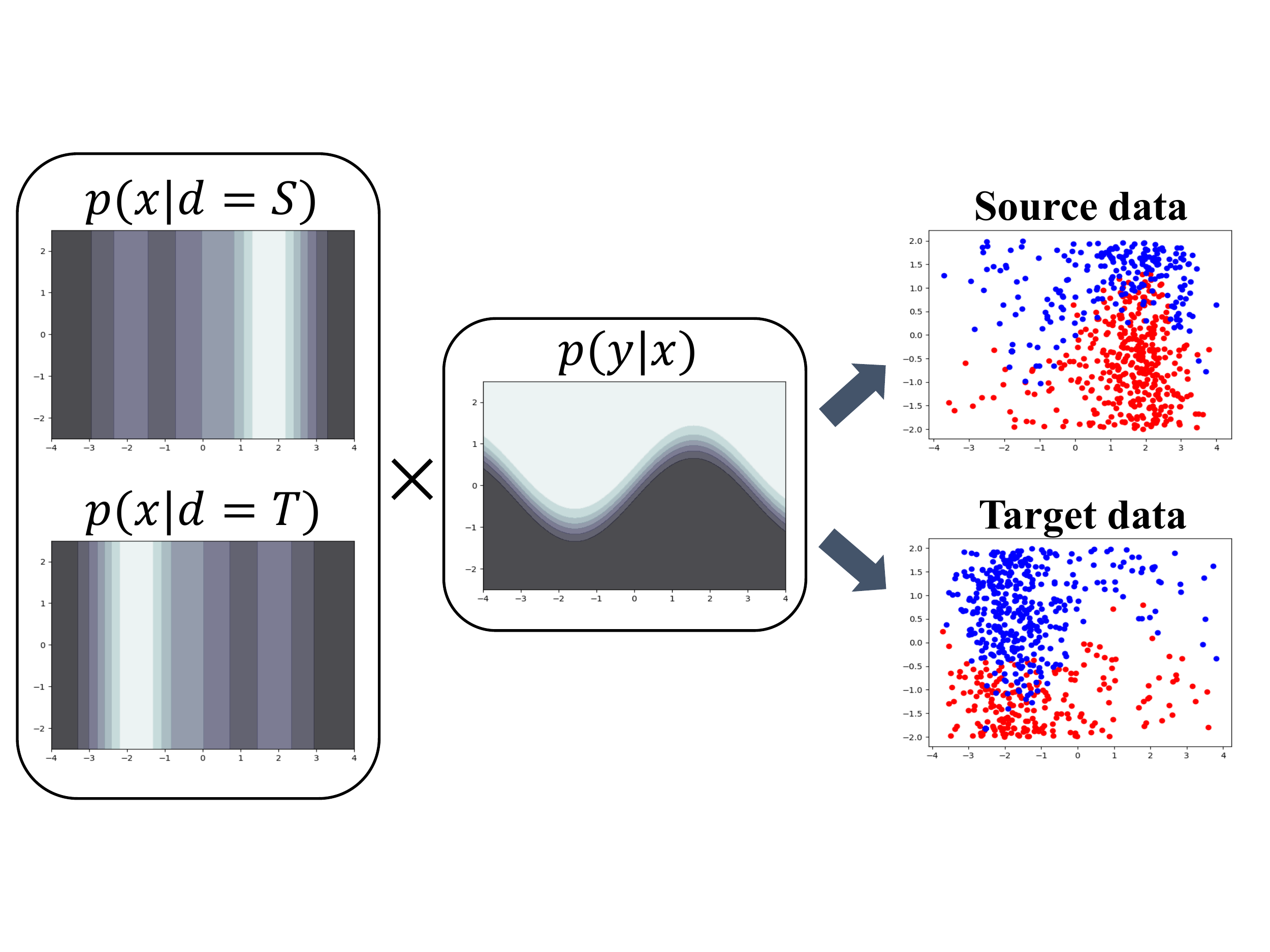}
\caption{Generation of toy datasets.}
\label{fig:toydata}
\end{figure}

To make the dataset, first, we generated the sample $(x, z)$ according to the data distribution that is previously described, then, we determined its label by randomly sampling according to the above posterior. Figure \ref{fig:toydata} shows a brief flow of how to generate the toy datasets. We generated 600 samples as source and target data, respectively. Note that we can obtain ground-truth $w(x)$ by calculating Eq. (\ref{eq:ratio}) with true probability density functions $p(x|d)$.

\begin{figure}[t]
\centering
\subcaptionbox{Without instance weights}{\includegraphics[scale=0.26]{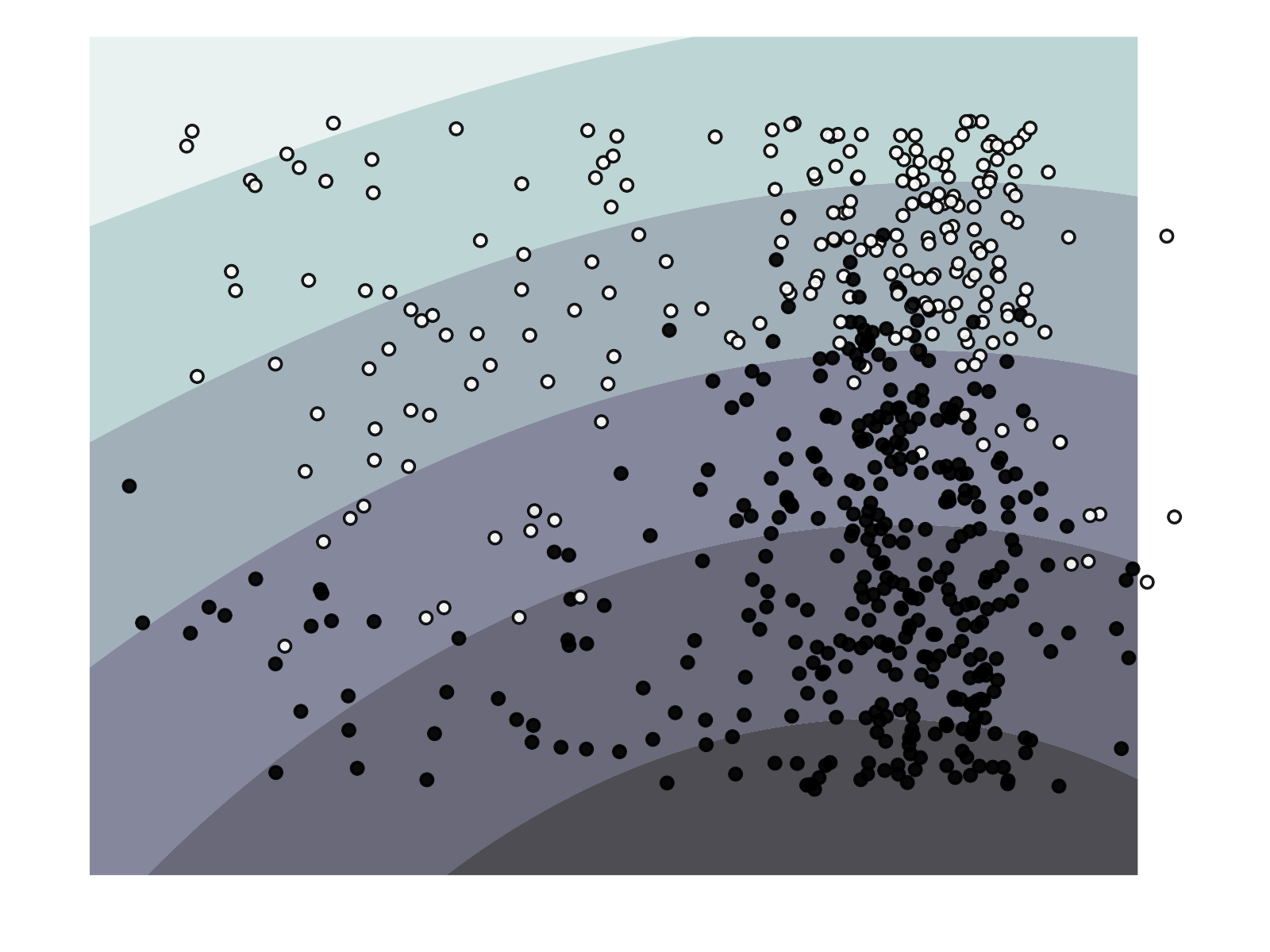}}
\subcaptionbox{With estimated weights}{\includegraphics[scale=0.26]{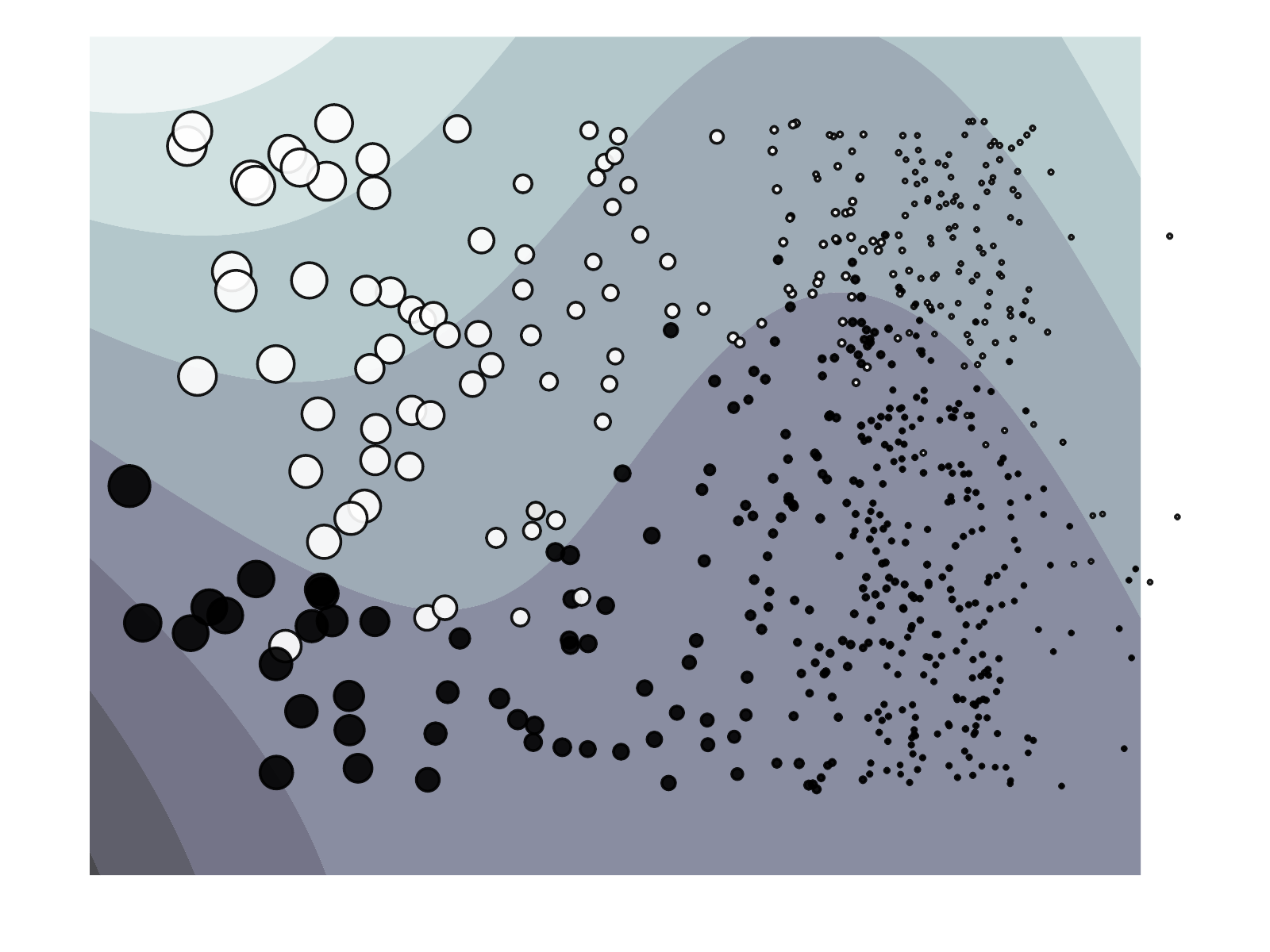}}
\caption{Visualization of instance weights and the trained classifier ($\circ$: positive-class instances, $\bullet$: negative-class instances).}
\label{fig:vis}
\end{figure}

First, we evaluated the accuracy of the weights estimated by our method by comparing them with the ground-truth weights. To quantitatively evaluate the accuracy, we compared our method with unconstrained Least-Squares Importance Fitting (uLSIF) \cite{Kanamori2009} that is one of the representative methods to estimate a probability density ratio. Using the target data, we estimated the weight by uLSIF, and compared its estimation error with that of our method. We measured the error by the root mean squared error. The results are shown in Table \ref{tab:err}. Although our method does not use any target data, it shows better performance than uLSIF. This indicates that attribute information can be more useful to estimate the probability density ratio. 

\begin{table}[tb]
\centering
\caption{The mixing ratios of GMM for toy datasets.}
\label{tab:GMM}
\begin{tabular}{|cc|ccccc|}
\hline
 \multicolumn{2}{|c|}{Dataset} & \multicolumn{5}{|c|}{Centroid} \\
 \multicolumn{2}{|c|}{} & -0.75$\pi$ & -0.5$\pi$ & 0.0 & 0.5$\pi$ & 0.75$\pi$ \\
\hline
A & $d=\mathrm{S}$ & 0.1 & 0.1 & 0.2 & 0.4 & 0.2 \\
 & $d=\mathrm{T}$ & 0.2 & 0.4 & 0.2 & 0.1 & 0.1 \\
\hline
B & $d=\mathrm{S}$ & 0.05 & 0.05 & 0.1 & 0.5 & 0.3 \\
 & $d=\mathrm{T}$ & 0.3 & 0.5 & 0.1 & 0.05 & 0.05 \\
\hline
C & $d=\mathrm{S}$ & 0.05 & 0.05 & 0.1 & 0.1 & 0.7 \\
 & $d=\mathrm{T}$ & 0.7 & 0.1 & 0.1 & 0.05 & 0.05 \\
\hline
\end{tabular}
\end{table}

\begin{table}[t]
\centering
\caption{The estimation error of weights.}
\label{tab:err}
\begin{tabular}{|c|ccc|}
\hline
 & \multicolumn{3}{|c|}{Dataset} \\
 & A & B & C \\
\hline
The proposed method & 0.179 & 0.573 & 0.679 \\
uLSIF & 0.291 & 0.664 & 0.743 \\
\hline
\end{tabular}
\end{table}

\begin{table}[t]
\centering
\caption{The accuracy of the trained SVM.}
\label{tab:err_SVM}
\begin{tabular}{|c|ccc|}
\hline
 & \multicolumn{3}{|c|}{Dataset} \\
 & A & B & C \\
\hline
 w/o weights & 91.3\% & 90.4\% & 88.1\% \\
 w/ estimated weights & 92.4\% & 91.0\% & 90.2\% \\
\hline
\hline
 w/ ground-truth weights & 92.4\% & 90.9\% & 90.4\% \\
\hline
\end{tabular}
\end{table}

We also evaluate the performance of our method as domain adaptation. We trained a classifier with weighted source data and tested it with the target data. To train a classifier, we used $C$-support vector machine ($C$-SVM) with the Gaussian kernel and tuned its hyper-parameters that are regularization coefficient $C$ and kernel width $\sigma$ by five-fold cross validation. We compared three methods: training without weights, training with estimated weights, and training with the ground-truth weights. Table \ref{tab:err_SVM} shows the accuracy of the SVM trained by each method. Our method achieved higher accuracy than that without importance weights and almost reached the same performance as that with ground-truth weights, though our method does not utilize ground-truth weights or any target data. Figure \ref{fig:vis} visualizes the instance weights and the trained classifier. The size of circles corresponds to the value of the instance weight, and contour lines represent the output of the decision function of SVM. Note that the true decision boundary is a sinusoidal curve as shown in Fig. \ref{fig:toydata}. Since only few source data are distributed at the left-hand side while many target data are at that side, large weights are assigned to those source data in our method, which results in a more accurate classifier especially at the left-hand side. 

\subsection{Experiments with benchmark dataset}

To evaluate our method in a more practical scenario, we conducted experiments with popular benchmark datasets on computer vision tasks. 

\subsubsection{MNIST dataset}

For the first experiment, we used the MNIST dataset \cite{LeCun1998} that contains handwritten digit images. The task is to classify these images into ten classes that correspond to digit numbers. We randomly chose 10,000 samples from the training data, and used them as source data, while the test data that includes 10,000 samples were used as target data. To make the source and target data have different data distributions, we clockwisely rotated each image with a randomly determined angle, where we set different probability distributions of the rotation angle for source and target data as shown in Table \ref{tab:ang}. We measured the performance of our method by the accuracy of the classifier trained with weighted source data similarly to the previous experiments. Instead of SVM, we used a deep neural network in this experiment. Table \ref{tab:net_MNIST} shows its network architecture that is loosely based on {\it LeNet} \cite{LeCun1998} but is modified by adding batch normalization layers. We trained the network by stochastic gradient descent with momentum, and the number of total update iterations was 10,000. To calculate the weight in our method, we estimated $p(z|x)$ by the $k$-nearest neighbor method with the features at the last hidden layer of the network. Since the calculation cost of the weight estimation is not small compared with that of the training network, we calculated the weights after each 100 iterations, and fixed them for the next 100 iterations. We used the weights to calculate the sampling probability of each sample when making a mini-batch. 

\begin{table}
\centering
\caption{The probability distributions of the rotation angle used in the experiment with the MNIST dataset.}
\label{tab:ang}
\begin{tabular}{|c|ccccc|}
\hline
 & \multicolumn{5}{|c|}{Rotation angle} \\
 & $-\frac{1}{3}\pi$ & $-\frac{1}{6}\pi$ & 0 & $+\frac{1}{6}\pi$ & $+\frac{1}{3}\pi$ \\
\hline
 Source & 0.05 & 0.05 & 0.1 & 0.5 & 0.3 \\
 Target & 0.3 & 0.5 & 0.1 & 0.05 & 0.05 \\
\hline
\end{tabular}
\end{table}

Table \ref{tab:result_MNIST} shows the accuracy of the trained classifier on the MNIST dataset. Without instance weights, the accuracy decreased from $97.1\%$ to $93.8\%$ when shifting from the source to target domains. On the other hand, our method suppressed this degradation of the classification performance, and achieved $94.9\%$ at the target domain. Interestingly, the accuracy at the source domain remains almost unchanged while adopting the instance weights. 

\begin{table}
\centering
\caption{The network architectures used in the experiments. MP2, BN, and FC denote $2 \times 2$ max-pooling, batch normalization, and a fully-connected layer, respectively.}
\label{tab:net}

\begin{minipage}{0.47\hsize}
\centering
\subcaption{MNIST}
\label{tab:net_MNIST}
\begin{tabular}{|c|c|}
\hline
 Layer type & \!\!\!\!\!\!\!\!
 \begin{tabular}{c}
  Size / num. \\
  of filters \\
 \end{tabular} \!\!\!\!\!\!\!\! \\
\hline
 conv. \!+\! ReLU & $5 \!\times\! 5$ \!/\! 20 \\
\hline
 MP2 \!+\! BN & $2 \!\times\! 2$ \!/\! 20 \\
\hline
 conv. \!+\! ReLU & $5 \!\times\! 5$ \!/\! 50 \\
\hline
 MP2 \!+\! BN & $2 \!\times\! 2$ \!/\! 50 \\
\hline
 FC \!+\! ReLU & 1 \!/\! 200 \\
\hline
 FC \!+\! softmax & 1 \!/\! 10 \\
\hline
\end{tabular}
\end{minipage}
\begin{minipage}{0.47\hsize}
\centering
\subcaption{Adience and VisDA2017}
\label{tab:net_face}
\begin{tabular}{|c|c|}
\hline
 Layer type & \!\!\!\!\!\!\!\!
 \begin{tabular}{c}
  Size / num. \\
  of filters \\
 \end{tabular} \!\!\!\!\!\!\!\! \\
\hline
 conv. \!+\! ReLU & $3 \!\times\! 3$ \!/\! 16 \\
\hline
 MP2 \!+\! BN & $2 \!\times\! 2$ \!/\! 16 \\
\hline
 conv. \!+\! ReLU & $3 \!\times\! 3$ \!/\! 24 \\
\hline
 MP2 \!+\! BN & $2 \!\times\! 2$ \!/\! 24 \\
\hline
 conv. \!+\! ReLU & $3 \!\times\! 3$ \!/\! 32 \\
\hline
 MP2 \!+\! BN & $2 \!\times\! 2$ \!/\! 32 \\
\hline
 FC \!+\! ReLU & 1 \!/\! 500 \\
\hline
 FC + softmax & 1 / 2 or 12 \\
\hline
\end{tabular}
\end{minipage}
\end{table}

\begin{table}[t]
\centering
\caption{Accuracy of the trained DNN on the MNIST dataset.}
\label{tab:result_MNIST}
\begin{tabular}{|c|cc|}
\hline
 & Target data & Source data \\
\hline
w/o weights & 93.8\% & 97.1\% \\
Our method & 94.9\% & 97.0\% \\
\hline
\end{tabular}
\end{table}

\subsubsection{Adience dataset}

For the second experiment, we used the Adience dataset \cite{Eidinger2014} that contains facial images with age and gender annotations. In this experiment, we conducted age estimation while considering gender as an attribute. Since eight age groups are defined in this dataset, age estimation can be formulated as an eight-class classification problem. There are five sub-datasets in this dataset, and we used the fifth sub-dataset as target data and the other sub-datasets as source data. While gender in this dataset is almost balanced, we artificially made it imbalanced in the target data to change the data distribution. We varied this imbalance, and evaluated our method for each setting. The network architecture for this experiment is shown in Table \ref{tab:net_face}. The number of total update iterations was 5,000. The other setting is the same as that in the previous experiment.

Table \ref{tab:result_face} shows the accuracy of the trained classifier on the Adience dataset. When the ratio between male and female samples in the target data is set to $[0.5, 0.5]$, the accuracy of our method is almost the same as that of the other methods. This is because the ratio in the source data is also balanced and the data distribution is almost the same between the source and target data. In contrast, when the ratio became imbalanced, our method achieved better performance. It indicates that the effectiveness of our method gets more significant as the discrepancy between the source and target data distributions becomes larger. The straightforward attribute-based weight did not lead to better performance, because it could not effectively utilize female samples in heavily imbalanced case. For example, when the ratio was set to $[0.9,0.1]$, the average weight of female examples was 9 times smaller than that of male examples in the straightforward method, while, in the proposed method, it became 2.2 times smaller, which is substantially more smooth weight than the straightforward method. 

\begin{table}[t]
\centering
\caption{Accuracy of the trained DNN on Adience dataset.}
\label{tab:result_face}
\begin{tabular}{|c|ccc|}
\hline
 & \multicolumn{3}{|c|}{[male, female] at target data} \\
 & $\![0.5, 0.5]\!$ & $\![0.7, 0.3]\!$ & $\![0.9, 0.1]\!$ \\
\hline
 w/o weights & 39.8\% & 40.0\% & 39.7\% \\
 \parbox[c][6ex]{9em}{The straightforward \\ attribute-based weight} & 39.3\% & 39.7\% & 39.9\%\\
 Our method & 39.9\% & 40.8\% & 41.4\% \\
\hline
\end{tabular}
\end{table}

\subsubsection{VisDA2017 dataset}

\begin{table}[t]
\centering
\caption{The number of data used in the experiment with the VisDA2017 dataset. $M$ was set to 24,000, and $r$ was varied in the experiment to control the discrepancy between domains.}
\label{tab:azi}
\begin{tabular}{|c|ccccc|}
\hline
 & \multicolumn{5}{|c|}{Azimuth of the captured objects} \\
 & \!10-61\! & \!78-129\! & \!146-197\! & \!214-265\! & \!282-333\! \\
\hline
 Source & $M$ & $M/r$ & $M/r$ & $M/r^2$ & $M/r$ \\
 Target & $M/r^2$ & $M/r$ & $M/r$ & $M$ & $M/r$ \\
\hline
\end{tabular}
\end{table}

\begin{table}[t]
\centering
\caption{Accuracy of the trained DNN on VisDA2017 dataset.}
\label{tab:result_visda}
\begin{tabular}{|c|ccc|}
\hline
 & \multicolumn{3}{|c|}{Dataset} \\
 & $\!r=2\!$ & $\!r=3\!$ & $\!r=4\!$ \\
\hline
 w/o weights & 95.6\% & 93.7\% & 91.5\% \\
 \parbox[c][6ex]{9em}{The straightforward \\ attribute-based weight} & 95.6\% & 93.7\% & 92.1\%\\
 Our method & 95.6\% & 94.0\% & 92.5\% \\
\hline
\end{tabular}
\end{table}

For more large-scale experiment, we used the VisDA2017 classification dataset \cite{VisDA2017}. This dataset contains object images with twelve categories, and the task is to discriminate the object category from the given image. Since the azimuth of the captured object is also provided in this dataset, we discretized the azimuth into five classes and used it as an attribute. We constructed the source and target data as shown in Table \ref{tab:azi}. Intuitively, the source domain is biased to ``front-view" images, while the target domain is biased to ``rear-view" images. We varied these bias by changing $r$ in Table \ref{tab:azi}. The network architecture and the setting for training the network are same as in the previous experiment.

Table \ref{tab:result_visda} shows the experimental result with VisDA2017 dataset. When $r$ is small, the discrepancy between the source and target domain is not large, which results in almost the same accuracy of all methods. As $r$ increases, the advantage of our method becomes large as same with the result of the previous experiment. 

\section{Conclusion}

In this paper, we proposed a zero-shot domain adaptation method based on attribute information. We showed how to estimate instance weights for source data by using the attribute information, and also clarified requirements for the attribute information to be useful, which is actually the same assumption adopted in some existing works. In addition, we revealed that our method can provide more precise estimation of sample-wise transferability than a straightforward attribute-based reweighting approach. Experimental results with both toy datasets and benchmark datasets showed that our method can accurately estimate the instance weights and performed well as domain adaptation. Future works include integration of our method with other recent domain adaptation methods and extension to the case in which the attribute information is partially available.

\bibliographystyle{unsrt}
\bibliography{myref}

\end{document}